\documentclass[11pt,a4paper]{article}
\usepackage{times,latexsym}
\usepackage[hyphens]{url}
\usepackage[T1]{fontenc}
\usepackage{amsfonts}

\usepackage[usenames,dvipsnames,svgnames,table]{xcolor}
\usepackage{colortbl}

\usepackage{multirow}
\usepackage{comment}
\usepackage{upquote}

\usepackage[acceptedWithA]{tacl2018v2}
\usepackage{xspace,mfirstuc,tabulary}

\newif\iftaclinstructions
\taclinstructionsfalse %
\iftaclinstructions
\renewcommand{\confidential}{}
\renewcommand{\anonsubtext}{(No author info supplied here, for consistency with
TACL-submission anonymization requirements)}
\newcommand{\instr}
\fi

\iftaclpubformat %

\else

\fi

\usepackage{tikz}
\usepackage{inconsolata}
\usepackage{enumitem}
\usepackage{tabu}
\usepackage{booktabs}
\usepackage{makecell}

\usepackage[disable]{todonotes}
\makeatletter
\newcommand*\iftodonotes{\if@todonotes@disabled\expandafter\@secondoftwo\else\expandafter\@firstoftwo\fi}  %
\makeatother
\newcommand{\noindentaftertodo}{\iftodonotes{\noindent}{}\ignorespaces}
\newcommand{\fixme}[2][]{{\todo[color=yellow,size=\scriptsize,fancyline,caption={},#1]{#2}}} %

\newcommand{\Fixme}[2][]{\fixme[inline,#1]{#2}\noindentaftertodo}

\newlength{\extramargin}
\usepackage{calc}
\iftodonotes{\usepackage[paperwidth=\paperwidth+\extramargin*2,marginparwidth=\marginparwidth+\extramargin,width=\textwidth,height=\textheight]{geometry}}{}

\newcommand{\dgraph}{dataflow graph\xspace}

\newcommand{\eg}{e.g.,\xspace}
\newcommand{\ie}{i.e.,\xspace}

\newcommand{\defn}[1]{\textbf{#1}}

\makeatletter
\renewcommand*\sectionautorefname{\S\@gobble}
\renewcommand*\subsectionautorefname{\S\@gobble}
\makeatother

\newsavebox{\DialogueBox}

\newenvironment{Dialogue}[1][\small]{
    #1
    
    \setlength\tabcolsep{7pt}
    \taburulecolor{lightgray}
    \vspace{.8em}
    \noindent
    \begin{tabu} to \columnwidth {|[2pt]lX}
}{
    \end{tabu}
    \vspace{.5em}
}

\newcommand{\User}[1]{User: & \UserUtt{#1} \\}
\newcommand{\AgentDo}[1]{
\\[-1.3em]
\multicolumn{2}{|[2pt]p{\linewidth}}{ 
\AgentAction{#1}
}
\\[.3em]
}
\newcommand{\AgentSay}[1]{Agent: & \AgentUtt{#1} \\}
\newcommand{\UserUtt}[1]{\textit{#1}}
\newcommand{\AgentUtt}[1]{\textit{#1}}
\newcommand{\AgentAction}[1]{\texttt{\small #1}}
\newcommand{\MetaAction}[1]{\texttt{\small \underline{#1}}}

\newcommand{\Salient}{\MetaAction{refer}\xspace}
\newcommand{\Revise}{\MetaAction{revise}\xspace}
\newcommand{\ReviseConstraint}{\MetaAction{reviseConstraint}\xspace}

\newcommand{\scSalient}{\texttt{\scriptsize\underline{refer}}\xspace}
\newcommand{\scRevise}{\texttt{\scriptsize\underline{revise}}\xspace}

\newcommand{\OurDataset}{SMCalFlow\xspace}

\title{\vspace{-1em} Task-Oriented Dialogue as Dataflow Synthesis \vspace{-1em}}

\newcommand{\authortab}[2]{
\begin{tabular*}{\textwidth}{c @{\extracolsep{\fill}} *{#1}{c}}
#2
\end{tabular*}
}

\author{%
  Microsoft Semantic Machines\thanks{\hspace{4pt}Please cite this paper as ``Semantic
  Machines et al. Task-Oriented Dialogue as Dataflow Synthesis. TACL 2020.''} ~~ \texttt{\textmd{<sminfo@microsoft.com>}} \\[.25em]
\bf\authortab{4}{Jacob Andreas & John Bufe & David Burkett & Charles Chen & Josh Clausman} \\
\bf \authortab{4}{Jean Crawford & Kate Crim & Jordan DeLoach & Leah Dorner & Jason Eisner} \\
\bf \authortab{5}{Hao Fang & Alan Guo & David Hall & Kristin Hayes & Kellie Hill & Diana Ho} \\
\bf \authortab{4}{Wendy Iwaszuk & Smriti Jha & Dan Klein & Jayant Krishnamurthy  & Theo Lanman} \\
\bf \authortab{3}{Percy Liang & Christopher H.\ Lin & Ilya Lintsbakh & Andy McGovern} \\
\bf \authortab{4}{Aleksandr Nisnevich & Adam Pauls & Dmitrij Petters & Brent Read & Dan Roth} \\
\bf \authortab{4}{Subhro Roy & Jesse Rusak & Beth Short & Div Slomin & Ben Snyder} \\
\bf \authortab{4}{Stephon Striplin & Yu Su & Zachary Tellman & Sam Thomson & Andrei Vorobev} \\
\bf \authortab{4}{Izabela Witoszko & Jason Wolfe & Abby Wray & Yuchen Zhang &
Alexander Zotov}\vspace{-1.5em}}

\date{}

\begin{document}
\maketitle
\begin{abstract}
  \vspace{-.8em}
  We describe an approach to task-oriented dialogue in which dialogue state is represented as a dataflow graph. A dialogue agent maps each user utterance to a program that extends this graph.  Programs include metacomputation operators for reference and revision that reuse dataflow fragments from previous turns. Our graph-based state enables the expression and manipulation of complex user intents, and explicit metacomputation makes these intents easier for learned models to predict. We introduce a new dataset, \OurDataset, featuring complex dialogues about events, weather, places, and people. Experiments show that dataflow graphs and metacomputation substantially improve representability and predictability in these natural dialogues. Additional experiments on the MultiWOZ dataset show that our dataflow representation enables an otherwise off-the-shelf sequence-to-sequence model to match the best existing task-specific state tracking model.
  The \OurDataset dataset, code for replicating experiments, 
  and a public leaderboard 
  are available at
  \url{https://www.microsoft.com/en-us/research/project/dataflow-based-dialogue-semantic-machines}.\looseness=-1
  \vspace{-1.5em}
\end{abstract}

\begin{figure}[t!]
    \centering
    \strut
    \includegraphics[width=.97\columnwidth,clip,trim=0in 0.6in 7in 0.1in]{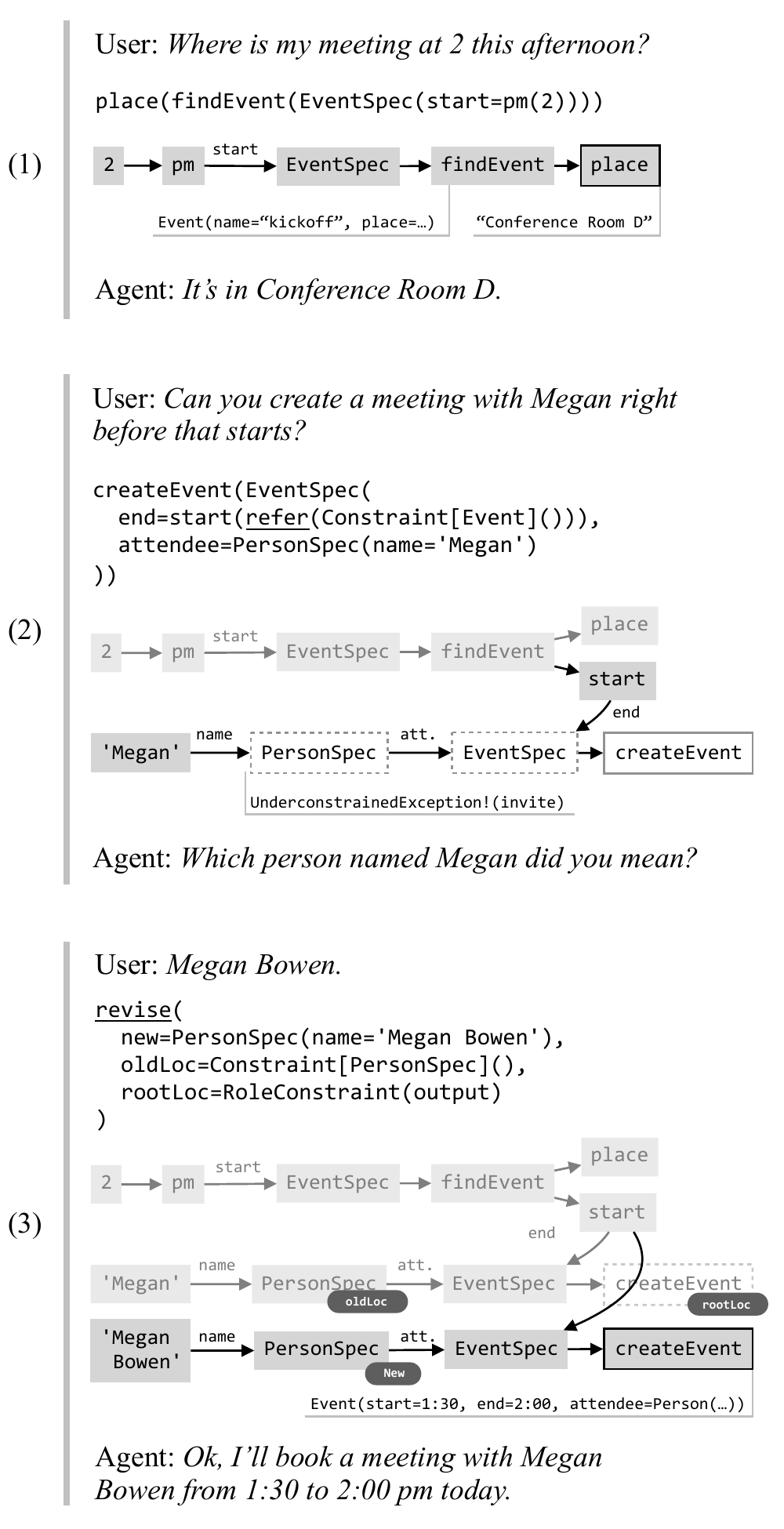}
    \vspace{-0.5em}
    \caption{%
        A dialogue and its dataflow graph. Turn (1) is an ordinary case of semantic parsing: the agent predicts a compositional query that encodes the user's question. 
        Evaluating this program produces an initial graph fragment.
        In turn (2), \textit{that} is used to refer to a salient \AgentAction{Event};
        the agent resolves it to the event retrieved in (1), then uses it in a subsequent computation.
        Turn (3)
        repairs an exception via a program that makes a modified copy of a graph fragment.
        \vspace{-2em}%
    }
    \label{fig:teaser}
\end{figure}

\section{Introduction}
\label{sec:intro}

Two central design decisions in modern conversational AI systems are the choices of state and action representations, which determine the scope of possible user requests and agent behaviors.
Dialogue systems with fixed symbolic state representations (like slot filling systems) are easy to train but hard to extend \cite{Pieraccini92Slots}. Deep continuous state representations are flexible enough to represent arbitrary properties of the dialogue history, but so unconstrained that training a neural dialogue policy ``end-to-end''
fails to learn appropriate latent states \cite{Bordes16EndToEndDialogue}.

This paper introduces a new framework for dialogue modeling that aims to combine the strengths of both approaches: structured enough to enable efficient learning, yet flexible enough to support
open-ended, compositional user goals that involve multiple tasks and domains.
The framework has two components: a new \textbf{state representation} in which dialogue states are represented as dataflow graphs; and a new \textbf{agent architecture} in which dialogue agents predict compositional programs that extend these graphs.
Over the course of a dialogue, a growing dataflow graph serves as a record of common ground: an executable description of the entities that were mentioned and the actions and computations that produced them (\autoref{fig:teaser}).

While this paper mostly focuses on representational questions, 
learning is a central motivation for our approach.  Learning to interpret natural-language requests is simpler when they are understood to specify graph-building operations.
Human speakers avoid repeating themselves in conversation by using
anaphora, ellipsis, and bridging to build on shared context \cite{Mitkov14Anaphora}.
Our framework treats these constructions by translating them into explicit \emph{metacomputation operators} for reference and revision, which directly retrieve fragments of the \dgraph that represents the shared dialogue state.
This approach borrows from corresponding ideas in the literature on program transformation 
\cite{Visser01ProgramTransformation}
and results in compact, predictable programs whose structure closely mirrors
user utterances.

Experiments show that our rich dialogue state representation
makes it possible to build better dialogue agents for challenging tasks.
First, we release a newly collected dataset of around 40K natural dialogues in English
about calendars, locations, people, and weather---the largest goal-oriented dialogue dataset to date. %
Each dialogue turn is annotated with a program implementing the user request. Many
turns involve more challenging predictions than traditional slot-filling,
with compositional actions, cross-domain interaction, complex
anaphora, and exception handling (\autoref{fig:smcalendar-example}).
On this dataset, explicit reference mechanisms
reduce the error rate of a seq2seq-with-copying model \cite{See2017PointerNet}
by 
5.9\% on all turns and by 
10.9\% on turns with a cross-turn reference.
To demonstrate breadth of applicability, we additionally describe how to automatically convert the simpler MultiWOZ
dataset into a dataflow representation.  This representation again enables a
basic seq2seq model to outperform a state-of-the-art, task-specific model at traditional state tracking.  
Our results show that within the dataflow framework,
a broad range of agent behaviors are both representable and learnable, and that
explicit abstractions for reference and revision are the keys to
effective modeling.

\section{Overview: Dialogue and Dataflow}
\label{sec:dataflow}

This section provides a high-level overview of our dialogue modeling framework, introducing
the main components of the approach. 
Sections~\ref{sec:salience}--\ref{sec:recovery} 
refine
this picture, describing the implementation and use of specific metacomputation operators.

We model a dialogue between a (human) \textbf{user} and an (automated)
\textbf{agent} as an interactive programming task where the human and computer
communicate using natural language.
Dialogue state is represented with a dataflow graph.
At each turn, the agent's goal is to translate the most recent user
utterance into a program.
Predicted programs nondestructively extend the dataflow graph,
construct any newly requested values or real-world side-effects, 
and finally describe the results to the user.
Our approach is significantly different from a conventional dialogue system pipeline, which has separate modules for language understanding, dialogue state tracking, and dialogue policy execution \cite{young2013ieee}.
Instead, a single learned model directly predicts executable agent 
actions and logs them in a graphical dialogue state.

\paragraph{Programs, graphs, and evaluation}

The simplest example of interactive program synthesis is \textbf{question answering}:

\begin{Dialogue}
    \User{When is the next retreat?}
    \AgentDo{%
      start(findEvent(EventSpec(\newline
      \strut~~name=\textquotesingle{}retreat\textquotesingle,\newline
      \strut~~start=after(now()))))
    }
    \AgentSay{It starts on April 27 at 9 am.}
\end{Dialogue}%

\noindent
Here the agent \defn{predicts} a program that invokes an API call (\AgentAction{findEvent}) on a structured input (\AgentAction{EventSpec}) to produce the desired query.\footnote{Note that what the agent predicts is not a formal representation of the utterance's \emph{meaning}, but a query that enables a contextually
appropriate \emph{response} (what \citet{Austin62How} called the ``perlocutionary force'' of the utterance on its hearer). The fact that \UserUtt{next} in this context triggered a search for ``events after now'' was learned from annotations. See \autoref{sec:data} for a discussion of how these annotations are standardized
in the \OurDataset dataset.}
This is a form of \textit{semantic parsing} \citep{Zelle95GeoQuery}.

The program predicted above can be rendered as a \textbf{\dgraph{}}:
\begin{center}
  \includegraphics[height=.9in,clip,trim=.12in 6.3in 8.1in 1.2in]{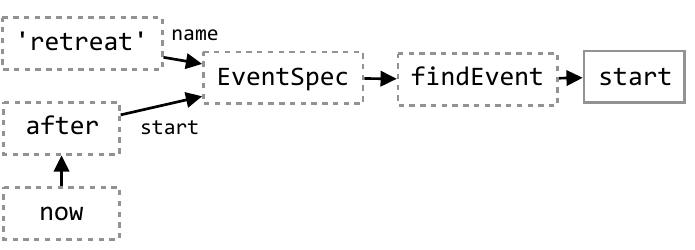} 
\end{center}
Each function call in the program corresponds to a node
labeled with that function. This node's parents correspond to the arguments of the function call. The top-level call that returns the program's result is depicted with a solid border.
A \dgraph is always acyclic, but is not necessarily a tree, as nodes may be reused.

Once nodes are added to a \dgraph, they are  \textbf{evaluated} in topological order.  Evaluating a node applies its function to its parents' values:
\begin{center}
  \includegraphics[height=.8in,clip,trim=.12in 6.4in 8.17in 1.25in]{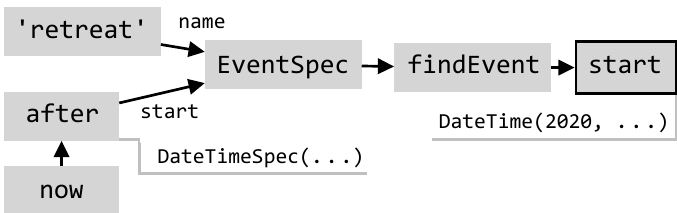} 
\end{center}
Here we have annotated two nodes to show that the value of \AgentAction{after(now())} is a \AgentAction{DateTimeSpec} 
and the value of the returned \AgentAction{start} node is a specific \AgentAction{DateTime}.  Evaluated nodes are shaded in our diagrams.  Exceptions (see \autoref{sec:recovery}) block evaluation, leaving downstream nodes unevaluated.

The above diagram saves space by summarizing the (structured) value of a node as a string.  In reality, each evaluated node has a dashed \defn{result edge} that points to the result of evaluating it:\looseness=-1
\begin{center}
  \includegraphics[height=.8in,clip,trim=.12in 6.2in 8.77in 1.35in]{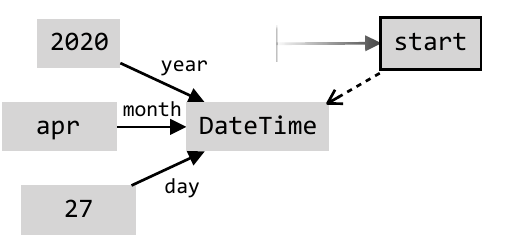} 
\end{center}
That result is itself a node in the dataflow graph---often a new node \emph{added} by evaluation. It may have its own result edge.\footnote{\label{fn:returncode}In other words, a function does not have to return a terminal node.  Its result may be an existing node, as we will see in \autoref{sec:salience}.  Or it may be a new non-terminal node, \ie the root of a subgraph that implements the function. The new nodes in the subgraph are then evaluated further, giving them result edges, although they also remain available for reference and revision.  Of course, a library function such as  \AgentAction{findEvent} or \AgentAction{+} that invokes an API will generally return its value directly as a terminal node.  However, translating natural language to higher-level function calls, which have been defined to expand into lower-level library calls (reminiscent of macro expansion), is often more easily learnable and more maintainable than translating it directly to the expanded graph.\looseness=-1} A node's \defn{value} is found by transitively following result edges until we arrive at a node whose result is itself.  Such a \defn{terminal node} is either a primitive value (\eg \AgentAction{2020}), or a constructor (\eg \AgentAction{DateTime}) whose parent nodes' values specify its arguments.  A constructor has the same (capitalized) name as the type it constructs.

\paragraph{Reference and revision} We now sketch two metacomputation functions whose evaluation extends the graph in complex ways.

As a representation of dialogue state, the dataflow graph records entities that
have been mentioned and the relationships among them.
All nodes in the \dgraph are eligible to be
referenced by subsequent utterances. 
Suppose, for example, that the user continues the previous dialogue fragment 
with a follow-up question:\looseness=-1

\begin{Dialogue}
    \User{What day of the week is that?}
    \AgentDo{dayOfWeek(\Salient{}())}
    \AgentSay{It's a Monday.}
\end{Dialogue}  

\noindent
The user's word \UserUtt{that} becomes the \Salient call in our predicted program, as it is a reference to ``some salient previously mentioned node.''  Evaluating \Salient here chooses the top-level node, \AgentAction{start}, from the previous turn. 
That node is then used as an argument to a  \AgentAction{dayOfWeek} node (nodes existing from previous turns are shown here in lighter ink):
\begin{center}
  \includegraphics[height=.9in,clip,trim=.12in 6.3in 8.05in 1.2in]{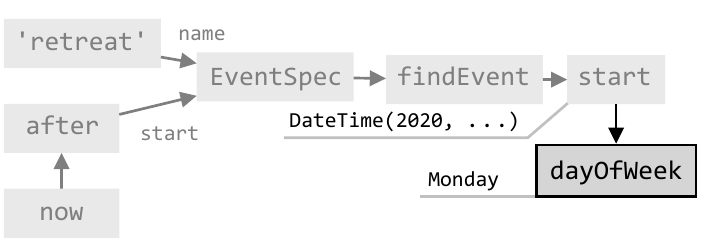}
\end{center}
and evaluating the latter node applies \AgentAction{dayOfWeek} to \AgentAction{start}'s value.
This diagram is actually a simplification: 
we will show in \autoref{sec:salience} how the \Salient call itself is also captured in the dataflow graph.

The user may next ask a question that changes the \emph{upstream} constraint on the event's start time:

\begin{Dialogue}
    \User{What about in 2021?}
    \AgentDo{%
    \Revise{}(\newline
    \strut~~new=DateTimeSpec(year=2021),\newline
    \strut~~oldLoc=Constraint[DateTimeSpec](),\newline
    \strut~~rootLoc=RoleConstraint(output))}
\end{Dialogue} \\[-1.5em]
A ``\AgentAction{new}'' \AgentAction{DateTimeSpec}
(representing \UserUtt{in 2021}) is to be substituted for some salient existing \AgentAction{old} node that has value type \AgentAction{DateTimeSpec} (in this case, the node \AgentAction{after(now())}).  The \Revise{} operator non-destructively splices in this new sub-computation and returns a revised version of the most salient computation containing \AgentAction{old} (in this case, the subgraph for the previous utterance, 
\AgentAction{root}ed at \AgentAction{dayOfWeek}):\looseness=-1

\begin{center}
    \includegraphics[height=.9in,clip,trim=.12in 6.1in 8.05in 1.4in]{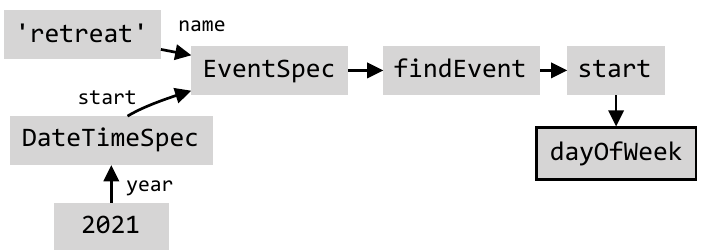}
\end{center}

As in the \Salient{} example, the target program (though not the 
above subgraph)
corresponds closely to the user's new utterance, making it easy to predict. Like the utterance itself, the program does not specify the revised subgraph in full, but describes how to find and reuse relevant structure from the previous \dgraph{}.

Given a dataset of turns expressed in terms of appropriate
graph-manipulation programs, the learning problem for a dataflow agent is the
same as for any other supervised contextual semantic parser.
We want to learn a function that maps user utterances to particular programs---a
well-studied task for which standard models exist.
Details of the model used for our experiments in this paper are
provided in \autoref{sec:experiments}.

\paragraph{Aside: response generation}

This paper focuses on \emph{language understanding}: mapping from a user's
natural language utterance to a formal response, in this case the value of the outlined
node returned by a program. Dialogue systems must also
perform \emph{language generation}: mapping from this formal response to a 
natural-language response. 
The dataset released with this paper includes output from a learned generation
model that can describe the value computed at a previous turn, describe the
structure of the computation that produced the value, and reference other nodes
in the dataflow graph via referring expressions. Support for structured,
computation-conditional generation models is another advantage of dataflow-based
dialogue state representations. While a complete description of
dataflow-based language generation is beyond the scope of this paper, we briefly describe the
components of the generation system relevant to the 
understanding system presented here.

The generation model is invoked after the evaluation phase. It conditions on a 
view of the graph rooted at the most recent return node, so generated responses
can mention both the previously returned value and the computation that produced it.
As the generation model produces the natural language response,
it \emph{extends the dataflow graph}. For example, if after the user query
``\emph{What's the date of the next retreat?}'' the agent responds:

\begin{Dialogue}
\AgentSay{It starts on April 27 at 9 am, and runs for 8 hours.}
\end{Dialogue}

\noindent
then it will also extend the dataflow graph to reflect that the event's duration was mentioned:

\begin{center}
    \includegraphics[height=.9in,clip,trim=.12in 6.1in 8.05in 1.4in]{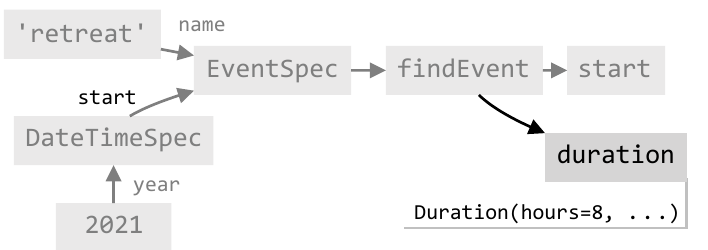}
\end{center}

\noindent
The duration of the event is now part of the
common ground in the conversation and available for future reference by either the
agent or the user. 

The generation model is also important for agent initiative:

\begin{Dialogue}
\User{Put an event on my calendar.}
\AgentSay{What should it be called?}
\end{Dialogue}

\noindent
As discussed in detail in \autoref{sec:recovery}, questions of this kind
can be generated in response to \emph{exceptions} generated by underspecified
user requests.
In the accompanying dataset release, the agent's
utterances are annotated with their dataflow graphs as extended by the generation model.
\Fixme{check}

\section{Reference resolution}
\label{sec:salience}

In a dialogue, entities that have been introduced once may be referred to again.
In dataflow dialogues, the entities available for reference are given by 
the nodes in the \dgraph{}. Entities are \textbf{salient} to conversation
participants to different degrees, and their relative salience determines the ways
in which they may be referenced \citep{lappin1994algorithm}.
For example, \textit{it} generally refers
to the most salient non-human
entity, while more specific expressions like
\textit{the Friday meeting} are needed to refer to accessible but less 
salient entities. Not all references to entities are overt: if the 
agent says ``\AgentUtt{You have a meeting tomorrow}'' and the user responds
``\UserUtt{What time?}'', the agent must predict the \emph{implicit} reference to a salient event.

\paragraph{Dataflow pointers}
We have seen that \Salient is used to find referents for referring expressions. In general, these referents may be existing dataflow nodes or new subgraphs for newly mentioned entities. We now give more detail about both cases. %

Imagine a dialogue in which the \dgraph{} contains the following fragment (which translates a mention of \UserUtt{Easter} or answers \UserUtt{When is Easter?}):
\begin{center}
    \includegraphics[height=0.3in,clip,trim=.12in 7.5in 9.1in .7in]{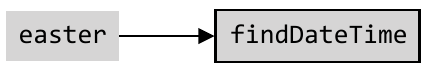}
\end{center}
\vspace{-.3em}
Suppose the user subsequently  mentions \emph{the day after that}. We wish to
produce this computation: \vspace{0.3em}
\begin{center}
    \includegraphics[height=.55in,clip,trim=.12in 6.9in 8.9in 1.1in]{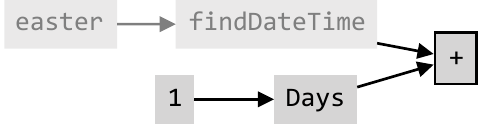}
\end{center}
In our framework, this is accomplished by mapping \UserUtt{the day after that} to
  \AgentAction{+(\Salient{}(), Days(1))}.
  The corresponding graph is not quite the one shown above, but it evaluates to the same value:
  
 \begin{center}
    \includegraphics[height=.9in,clip,trim=.12in 6.45in 8.4in 1.2in]{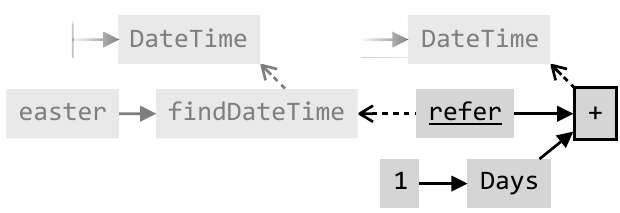}
 \end{center}
This shows how the \AgentAction{\Salient{}()} call is reified as a node in the \dgraph.  Its result is the salient \AgentAction{findDateTime} node from the previous turn---whose own result, a specific \AgentAction{DateTime}, now serves as the \emph{value} of \Salient{}.  We show both result edges here.  Evaluating \AgentAction{+} adds a day to this old \AgentAction{DateTime} value to get the result of \AgentAction{+}, a new \AgentAction{DateTime}.\looseness=-1

To enable \dgraph{} manipulation with referring expressions, all that is required is an implementation of \Salient that can produce appropriate pointers for both simple
references (\emph{that}) and complex ones (\emph{the first meeting}).

\paragraph{Constraints}

A call to \Salient{} is essentially a query that retrieves a node from the dialogue history, using a salience model discussed below. %
\Salient takes an optional argument: a \defn{constraint} on the returned node.
Indeed, the proper translation of \emph{that} in the context \UserUtt{the day after that} would be \AgentAction{\Salient{}(Constraint[DateTime])}.\footnote{Fortunately, this constraint need not be manually annotated.  Given the rest of the program, it can be inferred automatically by Hindley-Milner type inference~\cite{hindley_principal_1969,milner_theory_1978}, which establishes that this \scSalient{} node must return a \AgentAction{DateTime} if the program is to type-check.}
Constraints are predicates built from boolean connectives and the constructions illustrated below:\footnote{\label{fn:constraint-alias}\scSalient and \scRevise are not the only operations that take constraints as arguments.  For example, constraint arguments to \AgentAction{\scriptsize{findEvent}} and \AgentAction{\scriptsize{createEvent}} specify what sort of event is to be retrieved or created.  In these other contexts, to avoid distracting the reader, this paper uses \AgentAction{\scriptsize EventSpec} as an alias for \AgentAction{\scriptsize Constraint[Event]}.  It similarly uses aliases \AgentAction{\scriptsize PersonSpec} and \AgentAction{\scriptsize DateTimeSpec}.  (Our dataset does not.)}
\begin{itemize}[noitemsep,topsep=0pt]
            \item 
                \textbf{Type constraints}: \emph{the
                meeting} maps to the call
                \AgentAction{\Salient{}(Constraint[Event]())},
                where the constraint matches all nodes
                with values of type \AgentAction{Event}. 
            \item 
                \textbf{Property constraints}:
                Assuming a structured \AgentAction{Event} type with a
                \AgentAction{date} property, and a \AgentAction{Date} type with a
                \AgentAction{weekday} property, \emph{the Thursday
                meeting} maps to this nested call:\\
                \AgentAction{\Salient{}(Constraint[Event](date=\\
                \strut~~Constraint[DateTime](weekday=thurs)))}

              \item
                \textbf{Role constraints}: A role constraint specifies a keyword and matches nodes that are \emph{used as} keyword arguments with that keyword. For example, \emph{the month} maps to \AgentAction{\Salient{}(RoleConstraint(month))} and resolves to the constant node \AgentAction{apr} in the dialogues in  \autoref{sec:dataflow}, since that node was used as a named argument \AgentAction{month=apr}. We further allow keyword paths that select arguments of arguments: \AgentAction{thurs} in the previous bullet would satisfy \AgentAction{\Salient{}(RoleConstraint([date,weekday]))}.
        
\end{itemize}

To interpret a natural language referring expression,
the \emph{program prediction} model only needs to translate it into a contextually appropriate constraint $C$.  \AgentAction{\Salient{}($C$)} is then evaluated using a separate \emph{salience retrieval model} that returns an appropriate node.
The following dialogue shows referring expressions in action:\looseness=-1

\begin{Dialogue}
    \User{What's happening this morning?}
    \AgentDo{findEvent(EventSpec(\newline
    \strut~~start=and(today(), during(morning())))))}
    \AgentSay{You have a planning meeting at 9 am.}
    \User{What do I have after that?}
    \AgentDo{findEvent(EventSpec(start=after(end(\newline
    \strut~~\Salient{}(Constraint[Event]())))))}
    \AgentSay{Your birthday lunch is at 12:30 pm.}
    \User{How many people are coming to the 9~am meeting?}
    \AgentDo{length(attendees(\newline 
    \strut~~\Salient{}(Constraint[Event](start=am(9)))))
    }
    \AgentSay{5 people will be at the planning meeting.}
\end{Dialogue}

\noindent Here \UserUtt{the 9 am meeting} refers to the one that is salient from the first response, not an arbitrary one.

\paragraph{Salience retrieval model}

The salience retrieval model returns the most salient node satisfying the 
underlying constraint. Our dataflow framework is agnostic to the implementation 
of this model. A sophisticated model could select nodes via a machine-learned
scoring function. In the experiments in this paper, however, we rank nodes using a 
hard-coded heuristic.

The heuristic chooses the root node $r$ of the previous user utterance, if it satisfies the constraint.  More generally, the heuristic prefers nodes to the extent that they can be reached from $r$ in a small number of steps, where a step may move from a node to one of its input nodes, from an evaluated node to its result node, or from the root of an utterance to the root of an adjacent (user or system) utterance. 
If no satisfying node is found in the past several utterances, the heuristic 
falls back to generating code (see \autoref{fn:returncode}) that will search harder for a satisfying salient entity, perhaps by querying a database. 
For example, our earlier \AgentAction{Constraint[Event](start=am(9))} may return the expression \AgentAction{findEvent(EventSpec(start=am(9)))} if no \emph{9~am meeting} has been mentioned recently,
and \AgentAction{Constraint[Person](name=`Adam')} may return \AgentAction{findPerson(PersonSpec(name=`Adam'))} if no \emph{Adam} has been mentioned.  (See \autoref{fn:constraint-alias}.)

\section{Revision}
\label{sec:revision}

Beyond referring to previously mentioned \emph{entities} (nodes), task-oriented
dialogues frequently refer to 
previously executed \emph{computations} (subgraphs). This is one of the major
advantages 
of representing the dialogue state as a \dgraph of computations, not just a set of potentially salient entities.

\begin{Dialogue}
    \User{What time on Tuesday is my planning meeting?}
    \AgentDo{start(findEvent(EventSpec(\newline
      \strut~~name=\textquotesingle{}planning\textquotesingle{},\newline
      \strut~~start=DateTimeSpec(weekday=tuesday))))
  }
    \AgentSay{You meet with Grace at noon.}
    \User{Sorry, I meant all-hands.}
    \AgentSay{Your all-hands meeting is at 2:30 pm.}
\end{Dialogue}

\noindent
The second user utterance asks for the computation from the first user
utterance to be repeated, but with \UserUtt{all-hands} in place of \UserUtt{planning}.
The expected result is still a time, even though the second utterance makes no
mention of time.  

In the dataflow framework, 
we invoke a \Revise operator to construct the revised computation:

\begin{Dialogue}
    \User{Sorry, I meant all-hands.}
    \AgentDo{\Revise{}(%
            rootLoc=RoleConstraint(output),\newline
    \strut~~oldLoc=Constraint[String](),\newline
    \strut~~new=\textquotesingle{}all-hands\textquotesingle{})
    }
\end{Dialogue}

\noindent
Again, the content of the program closely reflects that of the corresponding utterance. The \Revise operator takes three arguments:
\begin{itemize}[nosep]
    \item \AgentAction{rootLoc}, a constraint to find the top-level node of the original computation;
    \item \AgentAction{oldLoc}, a constraint on the node to replace within the original computation;
    \item \AgentAction{new}, a new graph fragment to substitute there.
\end{itemize}
The \Revise node evaluates to the root of a \emph{modified copy} of the original computation, in which \AgentAction{new} now fills the role at the ``old'' location. Normally \AgentAction{rootLoc} is \AgentAction{RoleConstraint(output)}, which selects a node returned by a previous user utterance; thus, we \Revise that entire original request.

Revision is non-destructive---%
no part of the dialogue history is lost, so entities computed
by the original \texttt{target} and its ancestors remain available for
later reference.  However, the copy
shares nodes with the original computation where possible,
to avoid introducing unnecessary duplicate nodes that would have to be
considered by \Salient.

For the example dialogue at the beginning of this section, the first turn
produces the light gray nodes below.
The second turn adds the darker gray nodes, which specify the desired revision.  
\begin{center}
  \includegraphics[height=1.3in,clip,trim=.12in 4.6in 7.9in 2.5in]{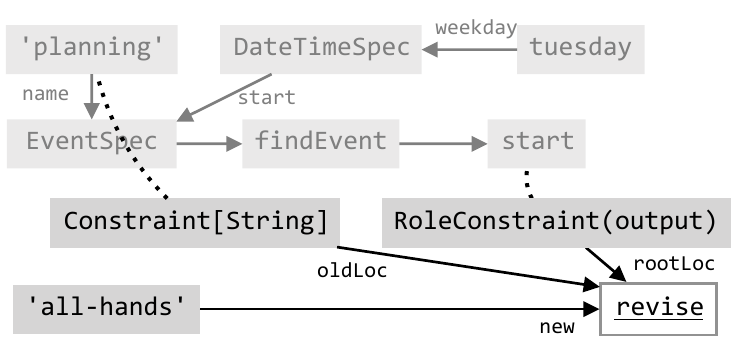}
\end{center}
Finally, evaluating the \Revise node selects the salient locations that match the \AgentAction{rootLoc} and \AgentAction{oldLoc} constraints (indicated in the above drawing by temporary dotted lines),
and constructs the revised subgraph (the new \AgentAction{start} node below and its ancestors).  The \emph{result} of evaluation (dashed arrow below) is the root of the revised subgraph.
\begin{center}
    \includegraphics[height=.85in,clip,trim=.12in 5.6in 8in 1.8in]{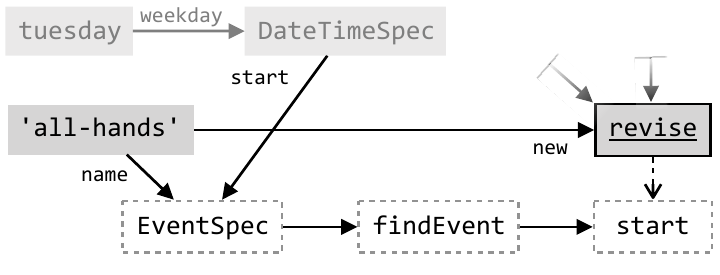}
\end{center}
Finally, evaluating these new nodes as well will establish that the value of the top-level \Revise is the start time of the `all-hands' meeting on Tuesday.

In the following example, the second utterance asks to replace a date specification.  However, the utterance appears in a context where the relevant \AgentAction{DateTimeSpec}-valued node to be ``replaced'' is an argument that has actually not yet been provided:

\begin{Dialogue}
  \User{When is lunch?}
  \AgentDo{start(findEvent(EventSpec(name=\textquotesingle{}lunch\textquotesingle{})))}
  \AgentSay{It's at 1 pm.}
  \User{What about tomorrow?}
  \AgentDo{\Revise{}(\newline
  \strut~~rootLoc=RoleConstraint(output),\newline
  \strut~~oldLoc=Constraint[DateTimeSpec],\newline
  \strut~~new=tomorrow())
  }
\end{Dialogue}

\noindent The revision replaces the missing \AgentAction{start} argument to the previous \AgentAction{EventSpec} (whose absence had resulted in some default behavior) with an \emph{explicit} argument (the \AgentAction{DateTimeSpec} returned by \AgentAction{tomorrow()}).  To achieve this, when the salience retrieval model is run with an \AgentAction{oldLoc} constraint, it must be able to return missing arguments that satisfy that constraint.  Missing arguments are implicitly present, with special value \AgentAction{missing} of the appropriate type.  In practice they are created on demand.%

\noindent
\begin{center}
    \includegraphics[height=1.2in,clip,trim=.6in 5in 7.25in 2in]{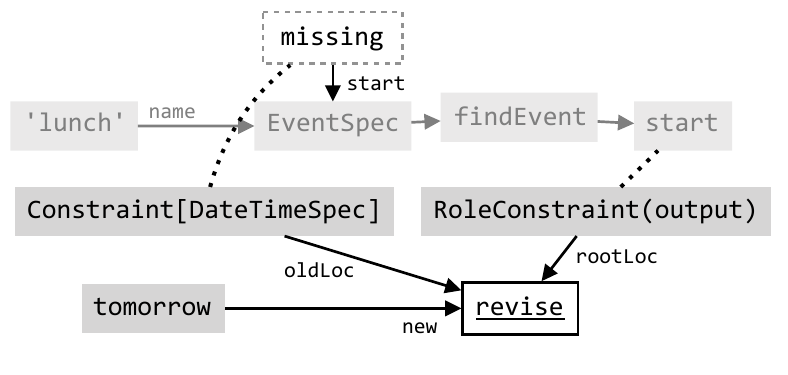}
\end{center}
Evaluating the \Revise node results in the new, revised subgraph pointed to by the dashed arrow, (which can then be evaluated):
\begin{center}
    \includegraphics[height=.8in,clip,trim=.2in 5.5in 7.7in 2in]{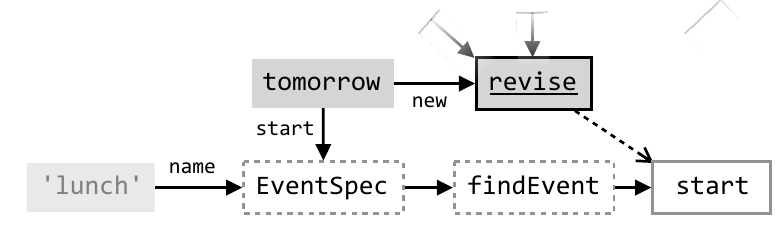}
    \vspace{-.5em}
\end{center}

Relatedly, a user utterance sometimes modifies a previously mentioned constraint such as an \AgentAction{EventSpec} (see \autoref{fn:constraint-alias}).  To permit this and more, we allow a more flexible version of \Revise to (non-destructively) transform the subgraph at \AgentAction{oldLoc} by applying a function, rather than by substituting a given subgraph \AgentAction{new}. Such functions are similar to rewrite rules in a term rewriting system \citep{Klop90TermRewriting}, with the \AgentAction{oldLoc} argument supplying the condition.
Our dataset (\autoref{sec:data}) makes heavy use of \ReviseConstraint calls, which modify a constraint as directed, while weakening it if necessary so that it remains satisfiable.  For example, if a 3:00--3:30 meeting is onscreen and the user says \emph{make it 45 minutes} or \emph{make it longer}, then the agent can no longer preserve previous constraints \AgentAction{start=3:00} and \AgentAction{end=3:30}; one must be dropped.

While the examples in this section involve a single update, real-world dialogues
(\autoref{sec:data})
can involve single user requests built up over
as many as five turns with unrelated intervening discussion.  Revisions of revisions or of constraints on reference are also seamlessly handled: \Revise can take another \Revise or a \Salient node as its \AgentAction{target}, leading to a longer chain of result edges (dashed lines) to follow.
Coordination of interactions among this many 
long-range dependencies remains a challenge even for modern attentional architectures \cite{Bordes16EndToEndDialogue}.
With \Revise all the needed information is in one
place; as experiments will show, this is crucial for good performance in more
challenging dialogues.

\section{Recovery}
\label{sec:recovery}

Sometimes users make requests that can be fulfilled only with the help of followup exchanges,
if at all. Requests might be incomplete:

\begin{Dialogue}
    \User{Book a meeting for me.}
    \AgentSay{When should the meeting start?}
\end{Dialogue}

\noindent
referentially ambiguous:

\begin{Dialogue}
    \User{Who is coming to the planning meeting?}
    \AgentSay{Susan Chu and Susan Brown.}
    \User{What is Susan's email?}
\end{Dialogue}

\noindent
or have no identifiable referent (a presupposition failure):

\begin{Dialogue}
    \User{When is my first meeting on February 30?} 
\end{Dialogue} \\[-2em]

\noindent Our solution is to treat such discourse failures as exceptions.  In principle, they are no different from other
real-world obstacles to fulfilling the user's request (server errors, declined
credit cards, and other business logic). To be useful, a dialogue model must
have some way to recover from all these exceptions, describing the problem to the user and guiding the dialogue past it. 

Our dialogue manager consists \emph{mainly} of an exception recovery mechanism.  This contrasts with traditional slot-filling systems, where a scripted policy
determines which questions to ask the user and in which order.
Scripted policies are straightforward
but cannot handle novel compositional utterances. 
Contextual semantic parsers treat compositionality, but 
provide no dialogue management mechanism at all.  Our dataflow-based approach allows the user to express complex compositional intents, but also allows the agent to reclaim the initiative when it is unable to make progress.  Specifically, the agent can elicit interactive repairs of the problematic user plan: the user communicates such repairs through the reference and revision mechanisms described in preceding sections.

\paragraph{Exceptions in execution}

In the \dgraph{} framework, failure to interpret a user utterance is signaled
by \textbf{exceptions}, which occur during evaluation.
The simplest exceptions result from
errors in function calls and constructors:

\begin{Dialogue}
    \User{What do I have on February 30?}
    \AgentDo{findEvent(EventSpec(\newline 
    \strut~~start=DateTimeSpec(month=feb, day=30)))
    }
\end{Dialogue}

\noindent
Evaluation of the \dgraph specified by this program cannot be completed. The \AgentAction{DateTimeSpec} constructor generates an exception,
and descendants of that node remain unevaluated. %

\begin{center}
    \includegraphics[height=.6in,clip,trim=.12in 6.6in 8in 1.3in]{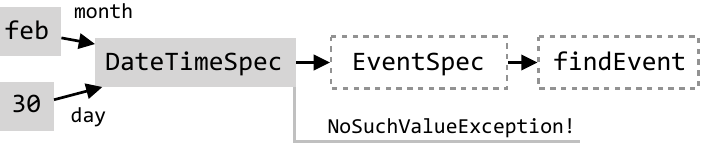}
\end{center}

An exception is essentially just a special result (possibly a structured value) returned by evaluation.  It appears in the dataflow graph, so the agent can condition on it when predicting programs in future turns. 
When an exception occurs, %
the generation model (\autoref{sec:dataflow}) is invoked on the exceptional node. This can be used to produce prompts like:

\begin{Dialogue}
    \AgentSay{There is no 30th of February. Did you mean some other date?}
\end{Dialogue}%

\noindent
At this point, recovering from the exception looks like any other revision step:
the user supplies a new value, and the agent simply needs to patch it into the
right location in the \dgraph. There are several answers the user could make,
indicating repairs at different locations:

\begin{Dialogue}
    \User{I meant February 28.}
    \vspace{-2pt}
    \AgentDo{\Revise{}(%
            rootLoc=RoleConstraint(output),\newline
    \strut~~oldLoc=Constraint[DateTimeSpec](),\newline
    \strut~~new=DateTimeSpec(month=feb, day=28))
    }
\end{Dialogue}

\vspace{-8pt}
\begin{Dialogue}
    \User{I meant March.}
    \vspace{-2pt}
    \AgentDo{\Revise{}(%
            rootLoc=RoleConstraint(output),\newline
    \strut~~oldLoc=Constraint[Month],\newline
    \strut~~new=mar)
    }
\end{Dialogue}

\noindent
The fact that exception recovery looks like any other turn-level prediction is
another key advantage of dataflow-based state representations. In the above
examples, the user specified a revision that would enable them to continue, but
they also would have been free to try another utterance (\UserUtt{List
all my meetings in February}) or to change goals altogether (\UserUtt{Never
mind, let's schedule a vacation}). 

Because of its flexibility, our exception-handling mechanism is suitable for
many situations that have not traditionally been regarded as exceptions. For
example, an interactive slot-filling workflow 
can be achieved via a sequence of 
underspecified constructors, each triggering an exception and eliciting a revision from the user:

\begin{Dialogue}
    \User{Create a meeting.}
    \AgentDo{%
    createEvent()\newline 
    $\dasharrow$ UnderconstrainedException!(name)
    }
    \AgentSay{What should it be called?}
    \User{Planning meeting.}
    \AgentDo{\Revise{}(%
            rootLoc=RoleConstraint(output),\newline
    \strut~~oldLoc=RoleConstraint(name),\newline
    \strut~~new=\textquotesingle{}Planning meeting\textquotesingle) \newline
    \strut $\dasharrow$ UnderconstrainedException!(start)}
    \AgentSay{When should it start?}
\end{Dialogue}

\noindent
The agent predicted that the user intended to revise the missing \texttt{name} because an exception involving the \AgentAction{name} path appeared in the dialogue history on the previous turn. %

Recovery behaviors are enabled by the phase separation between constructing the \dgraph (which is the job of program synthesis from natural language) and evaluating its nodes.  The \dgraph always 
contains a record of the user's current goal, even when the goal could not be successfully evaluated. This goal persists across turns and remains accessible
to reference, and thus can be interactively refined and clarified
using the same metacomputation operations as user-initiated revision.
Exception handling influences the course of the dialogue, 
without requiring a traditional hand-written or learned ``dialogue policy'' that reasons about full dialogue states.  Our policy only needs to generate language (recall \autoref{sec:dataflow}) that reacts appropriately to any exception or exceptions in the evaluation of the most recent utterance's program, just as it reacts to the ordinary return value in the case where evaluation succeeds.

\section{Data}
\label{sec:data}

\begin{figure}[t!]
{
\begin{Dialogue}[\scriptsize]
    \User{Can you remind me to go to the airport tomorrow morning at 8am?}
    \AgentDo{\scriptsize createCommitEventWrapper(\newline
    \strut~~createPreflightEventWrapper(\newline  
    \strut~~~~EventBuilder(\newline 
    \strut~~~~~~subject=\textquotesingle{}go to the airport\textquotesingle{},\newline 
    \strut~~~~~~start=dateAtTime(\newline
    \strut~~~~~~~~~date=tomorrow(),\newline
    \strut~~~~~~~~~time=numberAM(8)))))
    }
    \AgentSay{Does this look right? [showing an event card with start at 8 am and the event name]}
    \User{Make the location LaGuardia Airport.}
    \AgentDo{\scriptsize \textsc{\underline{reviseConstraint}}( \newline
    \strut~~rootLoc=RoleConstraint(output), \newline
    \strut~~oldLoc=Constraint[EventBuilder](),\newline
    \strut~~new=EventBuilder(\newline
    \strut~~~~location=LocationKeyphrase(\newline
    \strut~~~~~~\textquotesingle{}LaGuardia Airport\textquotesingle{})))
    }
    \AgentSay{How is this? [showing an updated card]}
    \User{What can I expect the weather there to be tomorrow morning?}
    \AgentDo{\scriptsize weatherQueryApi(\newline
    \strut~~place=findPlace(PlaceSpec(\newline
    \strut~~~~keyphrase=\scSalient{}(\newline
    \strut~~~~~~Constraint[LocationKeyphrase]()))),\newline 
    \strut~~time=DateTimeSpec(\newline
    \strut~~~~date=tomorrow(),\newline
    \strut~~~~constraint=morning()))
    }
    \AgentSay{It will be partly cloudy with a high of 77 °F and a low of 67 °F in LaGuardia Airport between 6:00 AM and 12:00 PM.}
    \User{How is the traffic on the way there?}
    \AgentDo{\scriptsize fenceNavigation()}
    \AgentSay{I can't answer questions about transit.}
    \User{That's fine. Book that please.}
    \AgentDo{\scriptsize confirmAndReturnAction()}
    \AgentSay{I've put that on your calendar. [showing the confirmed event card]}
\end{Dialogue}
}
\vspace{-1em}
\caption{A sample annotated dialogue in \OurDataset. 
Turn 1 features free-text subject and date/time.
Turn 2 features \ReviseConstraint.
Turn 3 features cross-domain interaction via \Salient and nested API calls (\AgentAction{findPlace} and \AgentAction{weatherQueryApi} are both real-world APIs).
Turn 4 features an out-of-scope utterance that is parried by a category-appropriate ``fencing'' response. 
Turn 5 confirms a proposal after intervening turns.
}
\label{fig:smcalendar-example}
\Fixme{are these examples consistent with our current dataset? (\autoref{fn:constraint-alias} already notes that PlaceSpec and DateTimeSpec are aliases for Constraints that are actually in the dataset)}
\end{figure}

To validate our approach, we crowdsourced a large English dialogue dataset, \OurDataset,
featuring task-oriented conversations about calendar events, weather, places, and people.
\autoref{fig:smcalendar-example} has an example. \OurDataset has several key characteristics:

\textbf{Richly annotated}:
    Agent responses are executable programs,
    featuring API calls, function composition, and complex constraints built from strings,
    numbers, dates and times in a variety of formats. They are not key-value structures
    or database queries, but instead full descriptions of the runtime behavior needed
    to react to the user in a real, grounded dialogue system.
    
\textbf{Open-ended}:
    We did not constrain crowdworkers to scripts.  Instead, they were given general information about agent capabilities and were encouraged to interact freely.
    A practical dialogue system must also recognize and respond to out-of-scope requests.
    Our dataset includes many such examples (see the fourth user turn in \autoref{fig:smcalendar-example}).

\textbf{Cross-domain}: 
    \OurDataset spans four major domains: calendar, weather, places, and people. 
    Cross-domain interaction is pervasive (\autoref{fig:smcalendar-example}).

To cover a rich set of back-end capabilities while encouraging worker creativity, we designed a wide range of \textit{scenarios} to guide dialogue construction. 
There are over 100 scenarios of varying topic and granularity.
Dialogues are collected via a Wizard-of-Oz process. Every dialogue is associated with a scenario. At each turn, a crowdworker acting as the user is presented with a dialogue as context and is asked to append a new utterance.
An annotator acting as the agent labels the utterance with a program (which may include \Salient and \Revise) and then selects a natural-language response from a set of candidates produced by the language generation model described in \autoref{sec:dataflow}. The annotation interface includes an autocomplete feature based on existing annotations. Annotators also 
populate databases of people and events 
to ensure that user requests have appropriate responses. %
The process is iterated %
for a set number of turns or until the annotator indicates the end of conversation.
A single dialogue may include turns from multiple crowdworkers and annotators. 

Annotators are provided with detailed guidelines containing example annotations
and information about available library functions.
Guidelines also specify conventions for pragmatic issues like the decision to annotate \emph{next} as \AgentAction{after} at the beginning of \autoref{sec:dataflow}. Crowdworkers are recruited from Amazon Mechanical Turk with qualification requirements such as living in the United States and with a work approval rate higher than 95\%.

Data is split into training, development, and test sets.
We review every dialogue in the \emph{test set} with two additional annotators. 75\% of turns pass through this double review process with no changes, which serves as an approximate measure of inter-annotator consensus on full programs.

\begin{table*}[t!]
    \centering
    \footnotesize
    \scalebox{0.94}{
    \begin{tabular}{@{}l cccccccc@{}}
    \toprule
         & \# Dialogues & \# User Turns &  \makecell{User Vocab. \\ Size} & \makecell{Library \\ Size} & \makecell{Utterance\\Length} & \makecell{Program\\Length} & \makecell{Program\\Depth} & OOS \\
         \midrule
         \OurDataset     & 41,517 & 155,923 & 17,397 & 338 & (5, 8, 11) & (11, 40, 57) & (5, 9, 11) & 10,466\\
         MultiWOZ 2.1    & 10,419 & 71,410 & 4,105 & 35 & (9, 13, 17) & (2, 29, 39) & (2, 6, 6) & 0 \\
    \bottomrule
    \end{tabular}
    }
    \caption{Dataset statistics. 
    ``Library\ Size'' counts distinct function names (\eg \AgentAction{findEvent}) plus keyword names (\eg \AgentAction{start=}).
    ``Length'' and ``Depth'' columns show (.25, .50, .75) quantiles. For programs, ``Length'' is the number of function calls and ``Depth'' is determined from a tree-based program representation.
    ``OOS'' counts the out-of-scope utterances.  MultiWOZ statistics were calculated after applying the data processing of \citet{Wu2019Trade}.
    Vocabulary size is less than reported by \citet{Goel2019HyST} because of differences in tokenization (see code release).\looseness=-1}
    \label{tab:data_statistics}
\end{table*}

For comparison, we also produce a version of the popular MultiWOZ 2.1 dataset \citep{Budzianowski18MultiWOZ,Eric2019MultiWOZ} with dataflow-based annotations.
MultiWOZ is a state tracking task, so in its original format the dataset annotates each turn with a dialogue state rather than an executable representation. 
To obtain an equivalent (program-based) representation for MultiWOZ, at each user turn
we automatically convert the annotation to a dataflow program.\footnote{We release the conversion script along with \OurDataset.}
Specifically, we represent each non-empty dialogue state as a call to an event booking function, \AgentAction{find}, whose argument is a \AgentAction{Constraint} that specifies the desired type of booking along with values for some of that type’s slots.  Within a dialogue, any turn that initiates a new type of booking is re-annotated as a call to \AgentAction{find}.  Turns that merely modify some of the slots are re-annotated as \ReviseConstraint calls.  Within either kind of call, any slot value that does not appear as a substring of the user’s \emph{current} utterance (all slot values in MultiWOZ are utterance substrings) is re-annotated as a call to \Salient with an appropriate type constraint, provided that the reference resolution heuristic would retrieve the correct string from earlier in the dataflow.  This covers references like \UserUtt{the same day}.  Otherwise, our re-annotation retains the literal string value. %

Data statistics are shown in \autoref{tab:data_statistics}.
To the best of our knowledge, \OurDataset is the largest annotated task-oriented dialogue dataset to date.
Compared to MultiWOZ, it features a larger user vocabulary, a more complex space of state-manipulation primitives, and a long tail of agent programs built from numerous function calls and deep composition.

\section{Experiments}
\label{sec:experiments}

We evaluate our approach on \OurDataset and MultiWOZ 2.1.
All experiments use the OpenNMT \cite{OpenNMT2017} 
pointer-generator network \cite{See2017PointerNet},
a sequence-to-sequence model that can copy tokens from the source sequence while decoding.
Our goal is to demonstrate that dataflow-based representations benefit standard neural model architectures.
Dataflow-specific modeling might improve on this baseline, and we leave this as a challenge for future work.

For each user turn $i$,
we linearize the target  program into a sequence of tokens $z_i$.
This must be predicted from the dialogue context---namely the concatenated source sequence
$x_{i-c}\,z_{i-c}\cdots x_{i-1}\,z_{i-1}\,x_i$ (for \OurDataset) or
$x_{i-c}\,y_{i-c}\cdots x_{i-1}\,y_{i-1}\,x_i$ (for MultiWOZ 2.1).
Here $c$ is a context window size, 
$x_j$ is the user utterance at user turn $j$, 
$y_j$ is the agent's natural-language response,
and $z_j$ is the linearized agent program.
Each sequence $x_j$, $y_j$, or $z_j$ begins with a separator token that indicates the speaker (user or agent).
Our formulation of context for MultiWOZ is standard \cite[\eg][]{Wu2019Trade}.
We take the source and target vocabularies to consist of all words that occur in (respectively) the source and target sequences in training data, as just defined.\looseness=-1

The model is trained using the Adam optimizer \cite{Kingma14Adam} with the maximum likelihood objective. 
We use 0.001 as the learning rate. Training ends when there have been two different epochs that increased the %
development loss.

We use Glove800B-300d (cased) and Glove6B-300d (uncased) \cite{Pennington2014Glove} to initialize the vocabulary embeddings for the \OurDataset and MultiWoZ experiments, respectively.
The context window size $c$,
hidden layer size $d$, 
number of hidden layers $l$,
and dropout rates $r$
are selected based on 
the agent action accuracy (for \OurDataset) 
or dialogue-level exact match (for MultiWoZ) 
on the 
development set
from \{2, 4, 10\}, \{256, 300, 320, 384\}, \{1, 2, 3\}, 
\{0.3, 0.5, 0.7\} 
respectively.
Approximate 1-best decoding uses a beam of size 5.

\begin{table}[t]
    \centering
    \footnotesize
    \scalebox{.82}{
    \begin{tabular}{l|cc|cc|cc}
    \toprule
         & \multicolumn{2}{c|}{Full}
         & \multicolumn{2}{c|}{Ref.~Turns} 
         & \multicolumn{2}{c}{Rev.~Turns} \\
         & dev & test & dev & test & dev  & test \\
         \midrule
         \# of Turns    & 13,499 & 21,224  & 3,554 & 8,965 & 1,052 & 3,315 \\
         \midrule
         Dataflow       & \bf .729 & \bf .665   & \bf .642 & \bf .574 & \bf .697 & \bf .565 \\
         inline         & .696 & .606   & .533 & .465 & .631 & .474 \\
    \bottomrule
    \end{tabular}
    }
    \caption{\textbf{\OurDataset} results. Agent action
    accuracy is significantly higher than a
    baseline without metacomputation, especially on turns that involve 
    a reference (Ref.\ Turns) or revision (Rev.\ Turns) to earlier turns in the dialogue ($p < 10^{-6}$, McNemar's test).}
    \label{tab:exp_smcal}
    \vspace{-7pt}
\end{table}

\paragraph{Quantitative evaluation} \autoref{tab:exp_smcal} shows results for the \OurDataset dataset.
We report program
accuracy: specifically, exact-match accuracy of the predicted program after inlining metacomputation (\ie replacing all calls to metacomputation operations with the concrete program fragments
they return).\footnote{%
    Specifically, we inline all \scSalient calls and \scRevise calls that involve direct substitution of the kind described in \autoref{sec:revision}. We preserve {\footnotesize\underline{\texttt{reviseConstraint}}} calls to avoid penalizing baselines that do not have access to pre-defined constraint transformation logic.
}
We also compare to baseline models that \emph{train} on inlined metacomputation. These experiments make it possible to evaluate the importance of explicit dataflow manipulation compared to a standard contextual semantic parsing approach to the task: a no-metacomputation baseline can still reuse computations from previous turns via the model's copy mechanism.

For the full representation, $c$, $d$, $l$, and $r$ are 2, 384, 2, and 0.5, respectively.
For the inline variant, they are 2, 384, 3, and 0.5. 
Turn-level exact match accuracy is around 73\% for the %
development set and 67\% for the test set.
Inlining metacomputation, which forces the model to explicitly resolve cross-turn computation,
reduces accuracy by 5.9\% overall, 10.9\% on turns involving references, and 9.1\% on turns involving revision. %
Dataflow-based metacomputation operations are thus essential for good model performance
in all three cases.

\begin{table}[t]
    \centering\footnotesize
    \scalebox{1}{
    \begin{tabular}{@{}l ccc@{}}
    \toprule
    & Joint Goal & Dialogue & Prefix \\
    \midrule
    Dataflow                & \bf .467  & \bf .220  & \bf 3.07 \\
    inline \scSalient       & .447      & \bf .202      & \bf 2.97 \\
    inline both             & \bf .467      & \bf .205      & 2.90 \\
    TRADE                   & .454      & .168      & 2.73 \\
    \bottomrule
    \end{tabular}
    }
    \caption{\textbf{MultiWOZ 2.1} test set results. 
    TRADE \cite{Wu2019Trade} results are from the public implementation.
    ``Joint Goal'' \citep{Budzianowski18MultiWOZ} is average dialogue state exact-match, ``Dialogue'' is average dialogue-level
    exact-match, and ``Prefix'' is the average number of
    turns before an incorrect prediction.  Within each column, the best result is boldfaced, along with all results that are not significantly worse ($p < 0.05$, paired permutation test).  Moreover, \emph{all} of 
    ``Dataflow,'' ``inline \Salient,'' and ``inline both'' have higher dialogue accuracy than 
    TRADE ($p < 0.005$).}
    \label{tab:exp_multwoz}
    \vspace{-2pt}
\end{table}

\begin{table}[t]
    \centering
    \footnotesize
    \begin{tabular}{@{}lr@{}}
      \toprule
        Error category  & Count \\
        \midrule
        \bf Underprediction & \bf 21 \\
        \midrule
        \bf Entity linking & \bf 21 \\
        ~~~~Hallucinated & 7 \\
        ~~~~Wrong type & 7 \\
        ~~~~Wrong field & 2 \\
        ~~~~Boundary mismatch & 5 \\
        \midrule
        \bf Fencing & \bf 22 \\
        ~~~~Should have fenced & 9 \\
        ~~~~Shouldn't have fenced & 8 \\
        ~~~~Wrong message & 5 \\
        \midrule
        \bf Ambiguity & \bf 23 \\
        ~~~~Wrong in context & 8 \\
        ~~~~Acceptable (same semantics) & 8 \\
        ~~~~Acceptable (different semantics) & 7 \\
        \midrule
        \bf Miscellaneous & \bf 10 \\
        ~~~~Used wrong function & 4 \\
        ~~~~Other / Multiple & 6 \\ %
        \midrule
        \bf Error in gold & \bf 3 \\
      \bottomrule
    \end{tabular}
    \caption{Manual classification of 100 model errors on the \OurDataset dataset. The largest categories are \emph{underprediction} (omitting steps from agent programs), \emph{entity linking} (errors in extraction of entities from user utterances, \emph{fencing} (classifying a user request as out-of-scope), and \emph{ambiguity} (user utterances with multiple possible interpretations). See \autoref{sec:experiments} for discussion.}
    \label{tab:error_analysis}
    \vspace{-1em}
\end{table}

We further evaluate our approach on 
dialogue state tracking using MultiWOZ 2.1.
\autoref{tab:exp_multwoz} shows results.
For the full representation, the selected model uses $c=2$, $d=384$, $l=2$, and $r=0.7$.
For the inline \Salient variant, they are 4, 320, 3, and 0.3.
For the variant that inlines both \Salient and \Revise calls,
they are 10, 320, 2, and 0.7.
Even without metacomputation, 
prediction of program-based 
representations gives results comparable to the existing state
of the art, TRADE, on the standard ``Joint Goal'' metric (turn-level exact match).
(Our dataflow representation for MultiWOZ is designed so that dataflow graph
evaluation produces native MultiWOZ slot-value structures.)
However, Joint Goal does not fully characterize the effectiveness of a state
tracking system in real-world interactions, as it 
allows the model to recover from an error at an earlier turn by 
conditioning on gold agent utterances after the error. We thus evaluate
on \emph{dialogue}-level exact match and prefix length (the average
number of turns until an error).
On these metrics the benefit of dataflow over past approaches is clearer. %
Differences within dataflow model variants are smaller here than in \autoref{tab:exp_smcal}.
For the Joint Goal metric, the no-metacomputation baseline is better; we attribute this to the comparative simplicity of reference in the MultiWOZ dataset.
In any case, casting the state-tracking problem as one of program prediction with appropriate
primitives gives a state-of-the-art
state-tracking model for MultiWOZ using only off-the-shelf sequence prediction tools.\footnote{%
A note on reproducibility: Dependence on internal libraries prevents us from releasing a full salience model implementation and inlining script for \OurDataset. The accompanying data release includes both inlined and non-inlined versions of the full dataset, and inlined and non-inlined versions of our model's test set predictions, enabling side-by-side comparisons and experiments with alternative representations. We provide full conversion scripts for MultiWOZ.
}

\paragraph{Error analysis}

Beyond the quantitative results shown in Tables \ref{tab:exp_smcal}--\ref{tab:exp_multwoz},
we manually analyzed 100 \OurDataset turns where our model mispredicted.
\autoref{tab:error_analysis} breaks down the errors by type.

Three categories involve straightforward parsing errors.
In \textbf{underprediction} errors, the model fails to predict some computation (\eg a search 
constraint or property extractor) specified in the user request. This behavior is not specific to 
our system: under-length predictions are also well-documented in neural machine translation systems \cite{Murray18NMTLength}.
In \textbf{entity linking} errors, the model correctly identifies the presence of an entity mention in the input utterance, but uses it incorrectly in the input plan. Sometimes the entity that appears in the plan is \emph{hallucinated}, appearing nowhere in the utterance; sometimes the entity is cast to a \emph{wrong type} (\eg locations interpreted as event names) used in the \emph{wrong field} or extracted with \emph{wrong boundaries}.
In \textbf{fencing} errors, the model interprets an out-of-scope user utterance as an interpretable command, or vice-versa (compare to \autoref{fig:smcalendar-example}, turn 4).

The fourth category, \textbf{ambiguity} errors, is more interesting. In these cases, the predicted plan corresponds to an interpretation of the user utterance that would be acceptable in \emph{some} discourse context. In a third of these cases, this interpretation is ruled out by either dialogue context (\eg interpreting \emph{what's next?} as a request for the next list item rather than the event with the next earliest start time) or commonsense knowledge (\emph{make it at 8} means 8 a.m.\ for a business meeting and 8 p.m.\ for a dance party). In the remaining cases, the predicted plan expresses an alternative computation that produces the same result, or an alternative interpretation that is also contextually appropriate.

\section{Related work}
\label{sec:related}

The view of dialogue as an interactive process of shared plan synthesis dates
back to \citeauthor{Grosz86Intentions}'s earliest work on discourse structure
(\citeyear{Grosz86Intentions, Grosz88Plans}). That work represents the state of a
dialogue as 
a predicate recognizing whether a desired piece of information has been
communicated or change in world state effected. 
Goals can be refined via questions and corrections from both users and 
agents. 
The only systems to attempt full versions of this shared-plans framework \citep[\eg][]{Allen96Robust,Rich01Collagen} required inputs that could be parsed under a predefined grammar. Subsequent
research on dialogue understanding has largely focused on two
simpler subtasks:

\textbf{Contextual semantic parsing} approaches focus on complex language
understanding without reasoning about underspecified goals or agent initiative.
Here the prototypical problem is iterated question answering
\citep{Hemphill90ATIS,Yu2019Sparc}, in which the user asks a sequence of
questions corresponding to database queries, and results of query execution are
presented as structured result sets. \citet{Vlachos14SP} describe a semantic
parsing representation targeted at more general dialogue problems.
Most existing methods interpret
context-dependent user questions (\emph{What is the next flight to Atlanta? When
does it land?}) by learning to copy subtrees
\citep{Zettlemoyer09ContextATIS,Iyyer2017SequenceQA,Suhr2018ContextATIS} or
tokens \citep{Zhang2019EditingSQL} from previously-generated queries.
In contrast, our approach reifies reuse with explicit graph 
operators.

\textbf{Slot-filling} approaches \citep{Pieraccini92Slots} model simpler utterances in the context of
full, interactive dialogues. 
It is assumed that any user intent can be
represented with a flat structure consisting of a categorical dialogue act
and a mapping between a fixed set of
slots and string-valued fillers. 
Existing fine-grained dialogue act schemes \citep{Stolcke2000DialogueActs} can distinguish among a
range of communicative intents not modeled by our approach,
and slot-filling representations have historically been
easier to predict \citep{Zue1994Pegasus} 
and annotate \citep{Byrne2019Taskmaster}.
But while recent variants support interaction between
related slots \cite{Budzianowski18MultiWOZ} and fixed-depth hierarchies of slots
\cite{Gupta18HierarchicalSlots}, modern slot-filling approaches remain 
limited in their support for semantic compositionality. By
contrast, our approach supports user requests corresponding to general
compositional programs.

More recent \textbf{end-to-end} dialogue agents attempt to map
directly from conversation histories to API calls and agent
utterances using neural sequence-to-sequence models without a 
representation of dialogue state
\citep{Bordes16EndToEndDialogue,Yu19CoSQL}. While
promising, models in these papers fail to outperform rule- or template-driven
baselines. \citet{Neelakantan2019NeuralAssistant} report greater success on a
generation-focused task, and promising results have also been obtained from hybrid neuro-symbolic dialogue systems \cite{Zhao2016Towards,williams2017hybrid,Wen2017EACL,Gao19ConvSurvey}. Much of this work is focused on improving agent modeling for existing representation schemes like slot filling. We expect that many modeling innovations (\eg the neural entity linking mechanism proposed by \citeauthor{williams2017hybrid}) could be used in conjunction with the new representational framework we have proposed in this paper.

Like slot-filling approaches, our framework is aimed at modeling full dialogues
in which agents can ask questions, recover from errors, and take actions with
side effects, all backed by an explicit state representation. However, our
notions of ``state'' and ``action'' are much richer than in slot-filling
systems, extending to arbitrary compositions of primitive operators. We use semantic parsing as a modeling framework for dialogue 
agents that can construct compositional states of this kind. While dataflow-based
representations are widely used to model execution state for programming
languages \cite{kam1976global}, this is the first work we are aware of that uses them to model conversational context and dialogue. 

\section{Conclusions}
\label{sec:conclusion}

We have presented a representational framework for task-oriented dialogue modeling based on
\dgraph{}s, in which dialogue agents predict a sequence of compositional updates
to a graphical state representation. 
This approach makes it possible to represent
and learn from complex, natural dialogues.
Future work might focus on improving prediction by introducing learned implementations of \Salient and \Revise that, along with the program predictor itself, could evaluate their hypotheses for syntactic, semantic, and pragmatic plausibility.
The representational framework could itself be extended, \eg by supporting
declarative user goals and preferences 
that
persist across utterances. %
We hope that the rich representations 
presented here---as well as our new dataset---will facilitate greater use of context and
compositionality in learned models for task-oriented dialogue.

\section*{Acknowledgments}

We would like to thank Tatsunori Hashimoto, Jianfeng Gao, and the
anonymous TACL reviewers for feedback on early drafts of this paper.

\bibliography{refs.bib}
\bibliographystyle{acl_natbib}

\end{document}